\documentclass[10pt,journal,compsoc]{IEEEtran}
\usepackage[utf8]{inputenc} % allow utf-8 input
\usepackage[T1]{fontenc}    % use 8-bit T1 fonts
\usepackage{url}            % simple URL typesetting
\usepackage{booktabs}       % professional-quality tables
\usepackage{amsfonts}       % blackboard math symbols
\usepackage{nicefrac}       % compact symbols for 1/2, etc.
\usepackage{microtype}      % microtypography
\usepackage[colorlinks,linkcolor=purple,citecolor=blue,urlcolor=blue]{hyperref} 

\usepackage{subcaption}

\usepackage{wrapfig,blindtext,lipsum}
\usepackage{graphicx}
\usepackage{amsmath}
\usepackage{bm}
\usepackage{balance}
\usepackage[export]{adjustbox}
\usepackage{xcolor}

\usepackage{algorithm,algpseudocode}
\usepackage{multirow}
\usepackage{multicol}
\usepackage{bbding}

\usepackage{authblk}

\newcommand{\etal}{\textit{et al.}}

\newcommand{\eg}{\textit{e.g.} }
\newcommand{\ie}{\textit{i.e.} }

\newcommand{\fig}{Figure}
\newcommand{\tab}{Table}
\newcommand{\eqn}{Equation}

\ifCLASSOPTIONcompsoc
  \usepackage[nocompress]{cite}
\else
  \usepackage{cite}
\fi

\begin{document}

% \title{Prior-Free Continual Learning with Unlabeled Data}
\title{Prior-Free Continual Learning with Unlabeled Data in the Wild}

\author{Tao Zhuo,
        Zhiyong Cheng,
        Hehe Fan,
        and~Mohan Kankanhalli,~\IEEEmembership{Fellow,~IEEE}
% \thanks{}
\thanks{This research is supported by the National Natural Science Foundation of China (No. 62002188), and Shandong Excellent Young Scientists Fund Program (Overseas) 2023HWYQ-114. (Corresponding author: Tao Zhuo).

Tao Zhuo, Zhiyong Cheng are with Shandong Artificial Intelligence Institute, Qilu University of Technology (Shandong Academy of Sciences), Jinan 250014, China, (e-mail: zhuotao724@gmail.com, jason.zy.cheng@gmail.com).

Hehe Fan is with Zhejiang University (hehe.fan.cs@gmail.com).

Mohan Kankanhalli is with the School of Computing, National University of Singapore, (e-mail: mohan@comp.nus.edu.sg).
}}

% \markboth{IEEE TRANSACTIONS ON PATTERN ANALYSIS AND MACHINE INTELLIGENCE}%
\markboth{}%
{Shell \MakeLowercase{\textit{et al.}}: Bare Advanced Demo of IEEEtran.cls for IEEE Computer Society Journals}

\IEEEtitleabstractindextext{
\begin{abstract}
Continual Learning (CL) aims to incrementally update a trained model on new tasks without forgetting the acquired knowledge of old ones. Existing CL methods usually reduce forgetting with task priors, \ie using task identity or a subset of previously seen samples for model training. However, these methods would be infeasible when such priors are unknown in real-world applications. To address this fundamental but seldom-studied problem, we propose a Prior-Free Continual Learning (PFCL) method, which learns new tasks without knowing the task identity or any previous data. First, based on a fixed single-head architecture, we eliminate the need for task identity to select the task-specific output head. Second, we employ a regularization-based strategy for consistent predictions between the new and old models, avoiding revisiting previous samples. However, using this strategy alone often performs poorly in class-incremental scenarios, particularly for a long sequence of tasks. By analyzing the effectiveness and limitations of conventional regularization-based methods, we propose enhancing model consistency with an auxiliary unlabeled dataset additionally. Moreover, since some auxiliary data may degrade the performance, we further develop a reliable sample selection strategy to obtain consistent performance improvement. Extensive experiments on multiple image classification benchmark datasets show that our PFCL method significantly mitigates forgetting in all three learning scenarios. Furthermore, when compared to the most recent rehearsal-based methods that replay a limited number of previous samples, PFCL achieves competitive accuracy. Our code is available at: {\color{magenta}https://github.com/visiontao/pfcl}.
\end{abstract}

}

\maketitle

\IEEEdisplaynontitleabstractindextext
\IEEEpeerreviewmaketitle

\section{Introduction}

\IEEEPARstart{H}{umans} are capable of acquiring new knowledge and skills over time without forgetting what they have previously learned. In contrast, conventional deep neural networks are often trained offline with the assumption that all data is available at once~\cite{NMI2022_Ven, TPAMI2017_Li, TPAMI2022_Boschini, arxiv2023_Zhou}. However, in dynamic environments, the model must incrementally learn new tasks. Due to privacy or storage concerns, directly updating a pre-trained model with only new datasets usually leads to drastic performance degradation on old tasks. This phenomenon is widely known as catastrophic forgetting~\cite{CS1995_Robins, ICCV2019_Lee, ECCV2020_Hayes}. To address this issue, Continual Learning (CL)~\cite{NMI2022_Ven} aims at preserving the learned knowledge of old tasks when learning new ones. 

According to different supervisory signals, there are three scenarios~\cite{NMI2022_Ven} for CL: Task-Incremental Learning ({\bf Task-IL}), Class-Incremental Learning ({\bf Class-IL}), and Domain-Incremental Learning ({\bf Domain-IL}). Both Task-IL and Class-IL learn new classes in streaming tasks. The difference is that task identity is available for Task-IL at both training and inference times, while it is unknown for Class-IL during inference. Therefore, Task-IL is easier than Class-IL, as it can select task-specific knowledge for each task with a given task identity. Domain-IL handles the data of different distributions or domain shifts, where the class spaces are kept the same. Besides, task identity is unknown at all times. 

\begin{figure}[t]
	\centering
	\includegraphics[width=\linewidth]{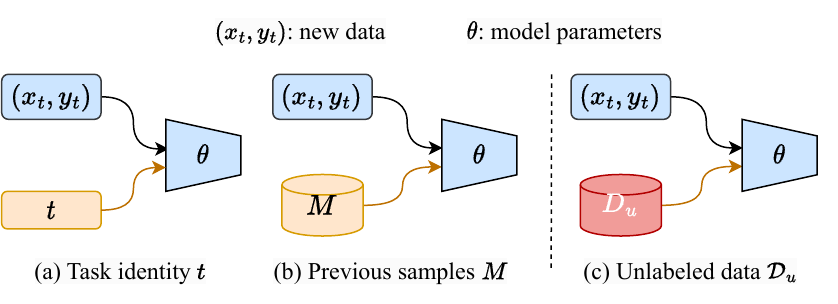}
	\caption{Main differences between the proposed {\bf Prior-Free Continual Learning (PFCL)} method and previous approaches. Compared to conventional methods that use task identity (a) or previous samples (b) during model training, our PFCL method (c) is more general and challenging due to the lack of task priors. Additionally, the unlabeled data used in PFCL can be collected in the wild without knowing the class labels of previous tasks, making it massive and free to obtain in practice.}
\label{fig_pfcl}
\vspace{-1em}
\end{figure}

Updating a pre-trained neural network on new tasks will inevitably overwrite the previously acquired knowledge, as the learned knowledge of a neural network is represented by its model parameters. Therefore, the core challenge in CL is to balance stability (preserving previous knowledge) and plasticity (learning new knowledge). Existing CL methods usually reduce forgetting by training the new model with additional task priors, such as task identity~\cite{Arxiv2016_Rusu, CVPR2021_Yan, CVPR2022_Wang} or previous samples~\cite{NIPS2017_Lopez, NIPS2019_Rolnick, CVPR2021_Bang, CVPR2022_Wang_vit}. Unfortunately, such priors might be unavailable in practice.

Task identity is a strong supervision signal for learning and distinguishing task-specific knowledge in CL. The early methods~\cite{Arxiv2016_Rusu, TPAMI2017_Li} often employ a multi-head architecture. The task identity must be required to select the correct output head at both training and inference times. To handle Class-IL scenarios, recent dynamic architectures~\cite{CVPR2022_Wang} usually employ task identity during the training stage only. By expanding the network capacity with a given task identity, the model can dynamically adapt to task-specific representations. Despite its effectiveness, task identity is unknown in Domain-IL scenarios and often unavailable in real-world applications. As a result, the strategy of reducing forgetting by using task identity becomes infeasible in these conditions. 

Another simple strategy to retain learned knowledge is to store a subset of previously seen samples in a memory buffer for rehearsal~\cite{CS1995_Robins, CVPR2017_Rebuffi, NIPS2019_Rolnick, CVPR2021_Bang, AAAI2023_Sarfraz, ICLR2022_AraniSZ}. However, when data privacy must be taken into account, storing raw data may not be allowed. Although some generative approaches~\cite{NIPS2017_Shin, ICCV2021_Smith} replace original samples with synthetic ones, generating high-quality data~\cite{NIPS2014_Goodfellow} suffers from potential risks of privacy leakage and requires expensive computation additionally. The performance of rehearsal-based methods heavily depends on the number of previous samples and often drops drastically when only a limited number of samples are available.

In this work, we propose a novel Prior-Free Continual Learning (PFCL) method that reduces forgetting without using task identity or any previous samples during training, see \fig~\ref{fig_pfcl}. Compared to previous methods~\cite{TPAMI2017_Li, TPAMI2022_Boschini, arxiv2023_Zhou}, the problem setting of our method is more general and challenging, \ie optimizing a pre-trained neural network on new tasks without forgetting. We address the prior-free CL from two aspects. (1) In contrast to previous methods that used multi-head architectures (\eg LwF~\cite{TPAMI2017_Li}) or dynamic networks (\eg L2P~\cite{CVPR2022_Wang}), we employ a fixed single-head architecture for all tasks, eliminating the need for task identity. (2) By using a regularization-based strategy that seeks consistent changes between the new and old models, we can avoid the requirement of revisiting previous samples. Ideally, if the output logits of a new model approximate its original ones, the forgetting issues would be alleviated. 

A significant limitation of the current regularization-based strategy is its inability to retain knowledge in Class-IL, especially when dealing with a long sequence of tasks~\cite{NIPS2020_Buzzega, NeuComp2021}. Based on extensive experiments, we empirically find that the effectiveness of using a regularization-based strategy alone depends on different tasks. In addition, it is worth mentioning that when a few previous samples are available for rehearsal, the regularization-based method even outperforms experience replay~\cite{ICLR2019_Riemer} sometimes, see the results reported in Section~\ref{sec_reg} and \tab~\ref{tbl_reg}. 

To further improve the performance of model regularization and consistently reduce forgetting, we propose to enhance model consistency with auxiliary datasets. As shown in the theoretical analysis in~\cite{ICML2020_Knoblauch}, regularization-based methods have an implicit and strong assumption on the geometry and nature of overlapping regions between the new model and the old one. However, this assumption is usually invalid in practice, leading to forgetting. To overcome this problem, we attempt to retain knowledge by increasing the overlapping regions in prediction spaces. Hence, we use a simple yet effective method that leverages auxiliary data to enhance model consistency across more data distributions. Since our experiments show that some auxiliary data may hurt performance, we further design a reliable sample selection method. Considering the motivation of increasing overlapping regions between model outputs, we filter out the data with low discrepancy measured by L1 distance. With the help of such a simple strategy, the robustness of the proposed PFCL method is significantly improved. 

Although the proposed PFCL method requires auxiliary data additionally, it is flexible to deploy. First, because the regularization loss does not need data labels, the auxiliary data used in our method can be unlabeled and freely collected in the wild in large quantities. Second, unlike rehearsal-based methods that store a set of previous samples in a memory buffer, the auxiliary data can be discarded after training. To verify the effectiveness of PFCL, we conduct extensive experiments on multiple image classification datasets. Moreover, we analyze the effects of different auxiliary datasets in our experiments. Evaluation results demonstrate that the proposed PFCL method significantly mitigates forgetting in all three CL scenarios and achieves remarkable performance. Furthermore, even compared to the most recent rehearsal-based methods~\cite{ICLR2022_AraniSZ, ICLR2023_Bhat, AAAI2023_Sarfraz} that replay a limited number of previous samples, the average accuracy of PFCL is competitive.

Our main contributions are summarized as follows.
\begin{enumerate}
    \item We propose a simple yet effective PFCL method that reduces forgetting without requiring task identity or previous samples for model training.
    \item We conduct a thorough analysis of the effectiveness and limitations of conventional regularization-based strategies in CL and propose to leverage auxiliary unlabeled data to assist model regularization.
    \item We develop a novel reliable sample selection method that consistently mitigates forgetting, as some auxiliary data may degrade the performance.
    \item Our experiments on multiple image classification datasets demonstrate that PFCL is effective in all three CL scenarios. Even when recent rehearsal-based approaches replay some samples, our PFCL method still achieves competitive performance. 
\end{enumerate}

\begin{figure*}[!t]
	\centering
	\includegraphics[width=0.7\linewidth]{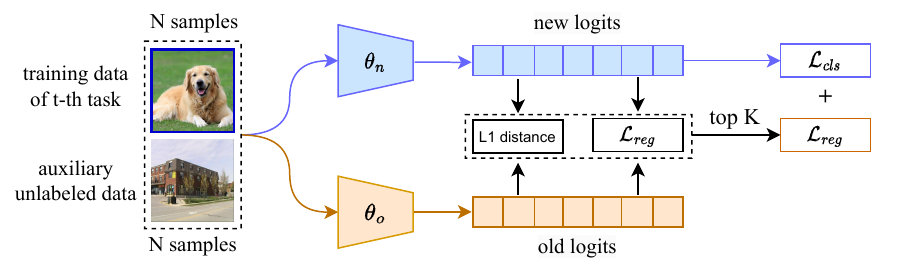}
	\caption{An overview of our PFCL method. $\theta_n$ denotes the parameters of the new model to be optimized while $\theta_o$ represents the parameters of the old model. $\mathcal{L}_{cls}$ is a classification loss for learning new knowledge and $\mathcal{L}_{reg}$ is a regularization loss for retaining previous knowledge. In addition, PFCL further mitigates forgetting by selecting top $K$ samples with high logit discrepancy measured by L1 distance.}
\label{fig_method}
\end{figure*}

\section{Related Work}

\subsection{Rehearsal-based Methods}
Rehearsal-based methods~\cite{CS1995_Robins, CVPR2017_Rebuffi, NIPS2017_Lopez, NIPS2019_Rolnick, CVPR2019_Hou, CVPR2021_Bang} reduce the forgetting issue by storing a subset of previously seen samples for joint training. For further performance improvement, recent rehearsal-based methods are often simultaneously used with other techniques, such as regularization~\cite{NIPS2020_Buzzega, ICCV2021_Cha, CVPR2022_Wang_vit, ICLR2022_AraniSZ, ICLR2022_Yoon, AAAI2023_Sarfraz}, dynamic architectures~\cite{ECCV2022_Foster, CVPR2022_Wang}, and meta learning~\cite{ICLR2019_Riemer}. Because storing raw data might be unavailable when data privacy has to be considered in some real-world applications, generative approaches~\cite{NIPS2017_Shin, ICCV2021_Smith} produce synthetic data with a deep generator (\eg GAN~\cite{NIPS2014_Goodfellow}). However, synthetic data still suffers from the potential risks of privacy leakage, and producing high-quality data is usually time-consuming. Another feature-replay strategy ~\cite{ECCV2020_Hayes, ECCV2020_Iscen, CVPR2020W_Liu, Arxiv2019_Pellegrini, NC2020_Ven} stores a subset of hidden representations (\eg CNN features). Despite avoiding raw data concerns and reducing the storage requirement, previously stored features would be unsuitable for the new model after learning new tasks. In addition, the performance of rehearsal-based approaches heavily relies on the number of available samples. It would drastically drop when available samples or storage resources are limited. More limits and merits of rehearsal-based methods are discussed in \cite{ICCV2021_Verwimp}. In this work, we focus on prior-free CL and the auxiliary data used in our method can be discarded after model training.

\subsection{Rehearsal-Free Methods}
Rehearsal-free methods do not use any raw or synthetic data from old tasks. We roughly divide current rehearsal-free methods mainly into two groups: dynamic architectures and regularization-based approaches. Some dynamic architectures~\cite{Arxiv2016_Rusu, CVPR2022_Wang, CVPR2023_Smith} are rehearsal-free and they do not store any previous samples. Their core idea is freezing some modules to preserve the knowledge of old tasks and expanding new trainable modules to learn knowledge of new tasks. These approaches usually require task identity to learn task-specific knowledge. The early method PNN~\cite{Arxiv2016_Rusu} requires task identity at both training and inference times, and thus it cannot be used in Class-IL. To overcome this drawback, recent methods (\eg L2P~\cite{CVPR2022_Wang} and CODA-Prompt~\cite{CVPR2023_Smith}) only use task identity to select learnable prompts during training. 

Regularization-based strategy is another rehearsal-free solution. By imposing a penalty term into the training loss, regularization-based methods attempt to reduce forgetting by seeking consistent model changes in parameter or prediction spaces. For example, EWC~\cite{PANS2017_kirkpatrick, ICML2018_Schwarz},  MAS~\cite{ECCV2018_Aljundi}, RW~\cite{ECCV2018_Chaudhry}, and SI~\cite{CVPR2019_Wu} prevent the model changes in network parameter spaces. LwF~\cite{TPAMI2017_Li} seeks consistency in prediction spaces. Besides, LwF employs knowledge distillation to reduce forgetting and use a given task identity to select the correct output head. Although much progress has been achieved by those approaches, recent studies~\cite{NeuComp2021, ICML2020_Knoblauch, NIPS2020_Buzzega} have shown that using a regularization-based strategy alone often performs poorly in Class-IL scenarios. In this work, we further study the strengths and limitations of model regularization in prediction spaces, and we propose to enhance model consistency by incorporating auxiliary unlabeled data. Task identity is unknown in Domain-IL scenarios and unavailable in many applications. Therefore, we use a fixed single-head architecture for all tasks, eliminating the need for task identity to select the correct output head.

\subsection{Continual Learning with Unlabeled Data}
\label{sec_unlabeled}
Unlabeled data has been used in several CL methods. 
Based on a given task identity, DMC~\cite{WACV2020_Zhang} first trains a separate model for new classes only. Then it employs an unlabeled dataset from a similar domain to combine the new model and the old one. GD~\cite{ICCV2019_Lee} utilizes a global knowledge distillation method on a sampled large-scale (1M) unlabeled dataset. Besides, GD further reduces forgetting by storing a subset of previous data for replay. When a few past samples are available, a recent method~\cite{CVPR2022_Tang} uses large-scale unlabeled data (\eg ImageNet) to generate diverse features that are semantically consistent with previous ones. Then it jointly trains the model on a subset of the old samples and auxiliary data. Bellitto \etal~\cite{ICPR2022_Bellitto} designed a rehearsal-based continual learning method that additionally leverages an auxiliary dataset for knowledge distillation. Compared to those methods, our method adopts a fixed single-head architecture and it does not need task identity or any previous samples during training. Thus the problem setting of our method is more challenging. Unlike the semi-supervised method Ordisco~\cite{CVPR2021_Wang}, which uses partially labeled data for continual learning, our approach employs a supervised learning strategy. Besides, Ordisco requires the unlabeled data to share the same labels as its training data, our method allows for the use of unlabeled data collected in the wild, without the need for shared labels.

\section{Prior-Free CL with Unlabeled Data}
We focus on a basic but often seldom-studied CL task that reduces forgetting without the knowledge of either task identity or previous samples. Based on a fixed single-head architecture, we leverage auxiliary unlabeled data to assist model regularization additionally. Moreover, the auxiliary data used in our method is not constrained by the class labels or domain distributions of previous tasks.

\fig~\ref{fig_method} shows an overview of the proposed PFCL framework. Given a mini-batch with $N$ new samples and $N$ auxiliary samples, we first compute the output logits on all samples with both the new model to be optimized and the old model. Then we measure the logit discrepancy between models with L1 distance and select the top $K$ samples with high discrepancy as reliable samples. Finally, a classification loss of new data and a regularization loss of reliable data are combined for model optimization. 

Next, we first describe the problem formulation of CL and the conventional regularization strategy with knowledge distillation. Then we study the effectiveness and limitations of the existing regularization approach with extensive experiments. Lastly, we introduce model regularization with auxiliary unlabeled data and a proposed reliable sample selection strategy.

\subsection{Problem Formulation}
Formally, we define $T$ tasks with corresponding datasets $\mathcal{D} = \{\mathcal{D}_1, \cdots, \mathcal{D}_T\}$ in a sequence, where $(x_t, y_t) \in \mathcal{D}_t$ denotes samples $x_t $ with ground truth labels $y_t$. Let $\theta$ be the model parameter, at a time step $t$, the goal of a CL  classification problem is to sequentially learn a function $f$ with optimized parameter $\theta$ on $(x_t, y_t)$ while maintaining the performance on previously seen data in $\{\mathcal{D}_1, \cdots, \mathcal{D}_{t-1}\}$. 

Without any CL techniques, the standard model fine-tuning on the $t$-th task can be achieved by minimizing a loss function $\mathcal{L}_{cls}$ as: 
\begin{equation}
    \mathcal{L}_{cls} = \mathbb{E}_{(x, y) \sim \mathcal{D}_t}[\ell_{ce}(f(x; \theta), y)],
    \label{eq_class}
\end{equation}
where $\ell_{ce}$ denotes the cross-entropy loss for multi-class classification, $f(x; \theta)$ represents the predicted logits of data $x$ with model parameter $\theta$. 

When all data is available for offline training, the loss function of conventional learning is represented as:
\begin{equation}
    \mathcal{L} = \sum_{t=1}^T \mathbb{E}_{(x, y) \sim \mathcal{D}_t}[\ell_{ce}(f(x; \theta), y)].
\end{equation}
However, the previous data might be inaccessible for joint training due to privacy or storage concerns. As a result, optimizing the new model on $\mathcal{D}_t$ with \eqn~\ref{eq_class} alone often performs poorly on $\{\mathcal{D}_1, \cdots, \mathcal{D}_{t-1}\}$, which is known as a catastrophic forgetting problem. We attempt to solve a prior-free continual learning task that reduces forgetting without knowing task identity and previous data. Due to the absence of task priors, the problem setting of our CL method is more challenging than that of previous approaches.

\subsection{Regularization with Knowledge Distillation}
Without revisiting any previous sample, we employ a regularization-based strategy to retain previous knowledge. Ideally, if an updated model has the same output logits as its original ones on the same data, it can be considered that the learned knowledge has not been forgotten. However, this condition cannot be satisfied in streaming tasks due to the presence of unseen classes. To address this issue, we approximate the new model's output logit distributions to its old ones by seeking consistent model changes with a penalty term. Based on such a regularization-based strategy, the forgetting issue can be alleviated.

Knowledge Distillation (KD)~\cite{Arxiv2015_Hinton} has been widely used as a regularization technique in CL. By imposing a penalty term into the loss function, the learned knowledge can be transferred from an old model (teacher) to a new one (student). Unlike the multi-head network architecture used in LwF~\cite{TPAMI2017_Li}, we employ a fixed single-head architecture as in DER++~\cite{NIPS2020_Buzzega}. Therefore, we directly use the total output spaces for all tasks after setting the maximum output dimensions. Compared to the typical method LwF, our method does not require task identity to select the correct output head for each task. 

Without loss of generality, the training loss of a knowledge distillation process can be formulated as:
\begin{equation}
     \mathcal{L} = \mathcal{L}_{cls} + \alpha \mathcal{L}_{reg},
    \label{eq_loss}
\end{equation}
where $\mathcal{L}_{reg}$ is a penalty term for model regularization and $\alpha > 0$ is a hyper-parameter balancing the trade-off between terms. Let $\theta_o$ be the parameters of the old model and $\theta_n$ be the parameters of the new model to be optimized. In prediction spaces, $\mathcal{L}_{reg}^t$ measures the consistency between two models on current training data $\mathcal{D}_t$ as:
\begin{equation}
    \mathcal{L}_{reg}^t = \mathbb{E}_{x \sim \mathcal{D}_t}[\ell_{dist}(f(x; \theta_n), f(x; \theta_o))].
    \label{eq_reg}
\end{equation}
In practice, $\ell_{dist}$ is usually computed by a Kullback-Leibler (KL) divergence loss.
% In practice, $\ell_{dist}$ is usually computed by a Kullback-Leibler (KL) divergence loss or a Mean Squared Error (MSE) loss. 

{\bf KL divergence loss.} Let $z=f(x; \theta)$ denote the output logits, the softened class probabilities $p^i$ of each class can be computed by using a temperature $\tau$ as: 
\begin{equation}
p^i=\frac{exp(z^i/\tau)}{\sum_j exp(z^j/\tau)}.
\end{equation}
Then the KL divergence loss between two logit distributions is computed as:
\begin{equation}
     \ell_{dist} = -\tau^2 \sum_{i=1} p_o^i \log(p_n^i),
    \label{eq_kl}
\end{equation}
where $i$ is the category index. A higher temperature $\tau$ produces softer probabilities over classes and provides a stronger signal for knowledge transfer. 

In our method, $\theta_n$ plays the role of plasticity and it aims to learn new knowledge by optimizing the model on $\mathcal{D}_t$. On the other hand, $\theta_o$ focuses on stability and it preserves previous knowledge with a regularization constraint $\mathcal{L}_{reg}^t$. By seeking consistent model changes during training, the regularization-based strategy attempts to balance stability and plasticity and then reduces forgetting.

\subsection{Effectiveness and Limitations of Regularization}
\label{sec_reg}
The data distributions of streaming tasks are usually non-i.i.d in real-world applications. Without any prior information about old tasks, reducing forgetting with a regularization-based method alone is very challenging. Previous studies~\cite{ICML2020_Knoblauch, NeuComp2021} have pointed out that using a conventional regularization method alone cannot achieve a decent performance in Class-IL scenarios, especially for a long stream of tasks. To well transfer the learned knowledge from an old model to a new one, we further study the effectiveness and limitations of knowledge distillation in Class-IL (disjoint class labels between tasks) with a series of experiments.

\begin{figure}[t]
\centering
\includegraphics[width=\linewidth]{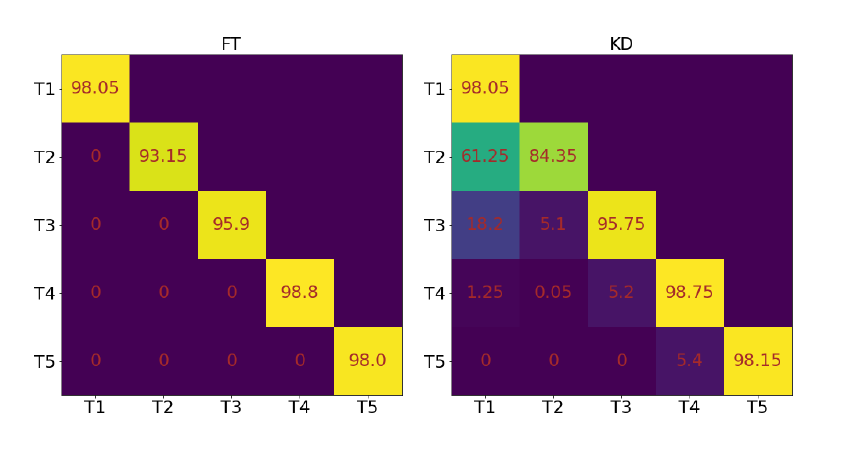} 
\caption{Average accuracy of FT (finetuning) and KD (knowledge distillation) after sequentially learning each task on the CIFAR10 dataset in Class-IL scenarios.}
\label{fig_kd}
\end{figure}

\begin{table}[t]
\caption{Average accuracy of Class-IL after training all tasks on CIFAR10 (5 tasks), CIFAR100 (5 and 10 tasks), and TinyImageNet (10 tasks). The backbone is ResNet18~\cite{NIPS2020_Buzzega} and the results are averaged across 3 runs. JT denotes the upper bound of jointly model training with all data. FT represents the lower bound of simple model finetuning. Besides, we set $\alpha=0.5$ for KD (knowledge distillation) in all experiments.}
\resizebox{\linewidth}{!}{
\begin{tabular}{ccccc} \toprule
Method                                  & CIFAR10-5 & CIFAR100-5 & CIFAR100-10 & TinyImg-10 \\  \midrule 
JT (upper bound)                        &  92.20  &  70.21    &   70.21   &   59.99    \\   
FT (lower bound)                        &  19.62  &  17.27     &    8.62   &   7.92    \\   \midrule
ER~\cite{ICLR2019_Riemer} (200 samples) &  44.79  &  21.94    & 14.23     &   8.49    \\
ER~\cite{ICLR2019_Riemer} (500 samples) &  57.74  &  28.02    & 21.54     &   9.99    \\  \midrule
KD                                      &  20.71  &  30.39    &   12.92    &  19.16   \\   \bottomrule
\end{tabular}}
\label{tbl_reg}
\end{table}

\fig~\ref{fig_kd} shows the average accuracy of KD and fine-tuning (FT) after sequentially learning each task on CIFAR10. It can be seen that KD effectively reduces forgetting for the second task. However, it fails to retain knowledge after training all 5 tasks and its performance drops significantly. Such results are the same as observed in previous studies~\cite{ICML2020_Knoblauch, NeuComp2021}. Additionally, we split CIFAR100 into 5 and 10 tasks, and TinyImageNet into 10 tasks. \tab~\ref{tbl_reg} reports the average accuracy in Class-IL after training all tasks. For comparison, \tab~\ref{tbl_reg} additionally presents the average accuracy of a typical rehearsal-based method Experience Replay (ER) with varying numbers of previous samples. Unlike the results observed on CIFAR10, KD  effectively reduces the forgetting issue on CIFAR100 and TinyImageNet. Moreover, it is worth mentioning that KD outperforms ER on CIFAR100-5 and TinyImg-10, even though ER replays 500 samples.  

Based on the above observations, we can conclude that model regularization can help to reduce forgetting in Class-IL, but its effectiveness relies on different tasks. Due to complex scenarios in dynamic environments, directly using such a regularization-based method is infeasible. To tackle this issue, we propose to enhance model consistency by distilling external knowledge from an auxiliary unlabeled dataset additionally.

\subsection{Reducing Forgetting with Auxiliary Unlabeled Data}
The core idea of model regularization in CL is to approximate the output logit distributions of the new model to its original ones. As the theoretical study in~\cite{ICML2020_Knoblauch}, regularization-based approaches have to make an implicit and strong assumption on the geometry and nature of overlapping regions between the new model and the old one. However, such an assumption is usually invalid in practice, leading to forgetting. Based on this analysis and experimental results observed in \tab~\ref{tbl_reg}, we raise a hypothesis that the forgetting issue can be alleviated by increasing prediction overlaps. 

To verify our hypothesis, we attempt to enhance prediction consistency between models. It is expected that seeking prediction consistency on more data distributions can further retain previous knowledge. To this end, we propose a simple strategy to increase the diversity of data distributions, \ie incorporating an auxiliary dataset. Similar to \eqn~\ref{eq_reg}, we distill external knowledge from an auxiliary dataset $\mathcal{D}_u$ as:
\begin{equation}
    \mathcal{L}_{reg}^u = \mathbb{E}_{x \sim \mathcal{D}_u}[\ell_{dist}(f(x; \theta_n), f(x; \theta_o))].
    \label{eq_unlabel}
\end{equation}
By seeking consistent predictions on both $\mathcal{D}_t$ and $\mathcal{D}_u$, the total loss is rewritten as:
\begin{equation}
    \mathcal{L} = \mathcal{L}_{cls} + \alpha (\mathcal{L}_{reg}^t + \mathcal{L}_{reg}^u).    
    \label{eq_all}
\end{equation}

Compared to most existing methods, our PFCL method requires auxiliary data to assist model regularization but it is still easy to deploy. First, the regularization loss does not need data labels (see \eqn~\ref{eq_unlabel}), and thus the auxiliary dataset can be unlabeled. Second, the unlabeled data used in our method is not constrained by the class labels or distributions of learned tasks, which makes it easy and free to collect. Finally, the auxiliary data can be discarded after model training and it does not occupy additional storage resources.

\begin{table}[t]
\centering
\caption{Average accuracy of Class-IL with different data for regularization. The experimental setup is the same as in \tab~\ref{tbl_reg}. In addition, we employ Caltech256~\cite{Caltech256} as the auxiliary dataset. }
\resizebox{\linewidth}{!}{
\begin{tabular}{ccccc} \toprule
Data                                & CIFAR10-5  & CIFAR100-5 & CIFAR100-10 & TinyImg-10 \\  \midrule 
$\mathcal{D}_t$                     &  20.71     &  30.39    &   12.92     &   19.16   \\
$\mathcal{D}_u$                     &  25.93     &  20.72    &   11.52     &   7.91   \\  
$\mathcal{D}_t \cup \mathcal{D}_u$  &  60.88     &  43.71    &   28.88     &   14.84  \\  \bottomrule
\end{tabular}}
\label{tbl_aux1}
\vspace{-1em}
\end{table}

\subsection{Reliable Sample Selection}
The purpose of using auxiliary unlabeled datasets is to enhance the model consistency on more data distributions. However, because of complex scenarios, some samples may hurt the performance. \tab~\ref{tbl_aux1} presents the accuracy of the regularization method with different datasets. It can be seen that incorporating an auxiliary dataset (Caltech256~\cite{Caltech256}) significantly reduces forgetting on CIFAR10-5, CIFAR100-5, and CIFAR100-10, but it degrades the performance on TinyImg-10. Therefore, directly using an auxiliary dataset for model regularization may cause negative impacts.

We propose a novel reliable sample selection method to achieve consistent performance improvement. It is important to note that regularization-based techniques aim to seek consistent predictions between two models. Samples with low discrepancies may not effectively increase the overlapping regions and may even degrade the model’s generalization. Therefore, we select a subset of reliable samples with large discrepancies. Specifically, we measure the discrepancy between two logit distributions with L1 metric as:
\begin{equation}
    d = |f(x; \theta_n)-f(x; \theta_o)|.
    \label{eq_rss}
\end{equation} 
Given $N$ current training samples and $N$ auxiliary samples in a mini-batch, we select $K$ ($K=N$ in our experiments) reliable samples with high discrepancy for regularization. Notice that we use a fixed $K$ for every mini-batch without knowing any priors of auxiliary data. Based on the proposed reliable sample selection method, the acquired knowledge is further maintained.

\section{Experiments}

% We first introduce the experimental setup and implementation details, and then we report the performance of our PFCL method in all three CL scenarios. We also provide extensive ablation study experiments to analyze the effectiveness of each component in PFCL. Finally, we discuss the effects of different auxiliary datasets.

\subsection{Experimental Setup}
{\bf Evaluation Datasets.} We evaluate PFCL on 4 image classification datasets. {\bf CIFAR10}~\cite{CIFAR2009_Krizhevsky} has 10 classes and each class consists of 5000 training images and 1000 test images of size $32 \times 32$. {\bf CIFAR100}~\cite{CIFAR2009_Krizhevsky} contains 100 classes and each class has 500 training images and 100 test images of size $32 \times 32$. {\bf Tiny ImageNet}~\cite{TinyImg2019_Chaudhry} has 200 classes and includes 100,000 training images and 10,000 test images of size $64 \times 64$ in total. {\bf Rotated MNIST}~\cite{NIPS2017_Lopez} is built upon the MNIST dataset~\cite{MNIST1998} and rotates the digits by a random angle in the interval $[0, \pi)$. In addition, Rotated MNIST has 60,000 training images and 10,000 test images of size $28 \times 28$. 

{\bf We mainly use Caltech256~\cite{Caltech256} as the auxiliary dataset} to assist model regularization, which consists of 30607 images. A detailed description of Caltech256 and more analysis of auxiliary datasets will be discussed in Section~\ref{sec_auxdata}. 

{\bf Evaluation metrics.} We evaluate the continual learning methods in terms of average accuracy and forgetting. Let $a_{T,t}$ denote the testing accuracy on $t$-th task when the model is trained on $T$-th task, and the final average accuracy after learning all $T$ tasks is computed as:
\begin{equation}
    Acc = \frac{1}{T}\sum_{t=1}^T a_{T,t}.
\end{equation} 
Besides, the average forgetting on $T$ tasks is computed as:
\begin{equation}
    Forget = \frac{1}{T-1}\sum_{t=1}^{T-1} max_{i \in \{1, \cdots, T-1\}}(a_{i,t} - a_{T,t}).
\end{equation} 

\subsection{Implementation Details}
{\bf Network architectures.} For a fair comparison, we use the same network architectures as in~\cite{NIPS2020_Buzzega}. Specifically, we employ ResNet18 without pre-training for CIFAR-10, CIFAR-100, and Tiny ImageNet. Besides, we utilize a fully-connected network with two hidden layers, each one having 100 ReLU units for the Roated MNIST dataset.

{\bf Model training.} Following the same settings in DER++\cite{NIPS2020_Buzzega, TPAMI2022_Boschini}, we use Stochastic Gradient Decent (SGD) as the optimizer. We train CIFAR10 and CIFAR100 with 50 epochs, Tiny ImageNet with 100 epochs, and Rotated MNIST with one epoch. The learning rate is 0.03 and the size of the mini-batch is 32 (64 for Rotated MNIST). In all experiments, $K$ is equal to the batch size. In addition, we define a set of epochs at which the learning rate is divided by 10 ([35; 45] for CIFAR-10 and CIFAR-100, [70; 90] for TinyImageNet).  For knowledge distillation, the temperature $\tau$ is set to 2 as in LwF~\cite{TPAMI2017_Li}. The balancing parameter is $\alpha=0.5$ on CIFAR10 and CIFAR100, $\alpha=1$ on TinyImageNet and Rotated MNIST. For data augmentations, we employ random cropping, horizontal flip, ColorJitter, and grayscale.

Since the images from the auxiliary dataset are of different sizes, we resize them to the same resolutions as that of the target dataset. Besides, because the images of Rotated MNIST are binary, we convert the auxiliary data into grayscale for t the same input dimensions. Additionally, as discussed in~\cite{ICCV2019_Cho}, KL divergence may dominate the loss at the end of training and hurt the overall accuracy of the student model. To address this issue, we stop knowledge distillation for the last 5 training batches of each task~\cite{ICCV2019_Cho}, which encourages the convergence of the student model on new tasks.

{\bf Batch normalization issue.} To speed up model training, Batch Normalization (BN) has been widely used in deep neural networks. However, the streaming data in the CL task is usually non-i.i.d, and thus the discrepancy between training and inference in BN severely hurts the performance on previous tasks. Instead of designing a new normalization mechanism~\cite{ICLR2022_Pham}, we use a simple way to solve this problem. Specifically, we concatenate current training data and auxiliary data in one mini-batch and feed them into the neural network together. Unlike feeding them separately, concatenating them in one mini-batch can make the input data with various distributions, improving the model generalization. Despite its simplicity, this strategy effectively solves the BN issue in our experiments.

\begin{table*}[!t]
\centering
\caption{Classification results of different CL models on three benchmark datasets, which is averaged over 3 runs. We report the average Top-1 (\%) accuracy of all tasks after training. Besides, we split CIFAR10 into 5 tasks and Tiny ImageNet into 10 tasks, Rotated MNIST has 20 tasks. ``-'' denotes the results are not reported in published papers. ``$*$'' indicates incompatibility issues, because of an unknown task identity in Domain-IL.}
\resizebox{0.85\linewidth}{!}{
\begin{tabular}{clccccc}
\toprule
\multirow{2}{*}{\begin{tabular}[c]{@{}l@{}}Prior  \end{tabular}} & \multirow{2}{*}{Method} & \multicolumn{2}{c}{CIFAR10-5}  & \multicolumn{2}{c}{TinyImg-10} & RMNIST-20    \\ 
&   & Class-IL    & Task-IL      & Class-IL     & Task-IL      & Domain-IL \\ \midrule
\multirow{2}{*}{-} 
& JT (upper bound)	&	92.20 $\pm$ 0.15	&	98.31 $\pm$ 0.12	&	70.56 $\pm$ 0.28	&	82.04 $\pm$ 0.10  & 95.76 $\pm$ 0.04\\  
& FT (lower bound)	&	19.62 $\pm$ 0.05	&	61.02 $\pm$ 3.33	&	7.92 $\pm$ 0.26	&	18.31 $\pm$ 0.68 &  67.66 $\pm$ 8.53 \\ \midrule  
\multirow{12}{*}{500 samples} 
& ER~\cite{ICLR2019_Riemer} 	&	 57.74 $\pm$ 0.27 	&	 93.61 $\pm$ 0.27 	&	 9.99 $\pm$ 0.29 	&	 48.64 $\pm$ 0.46	& 88.91 $\pm$ 1.44	\\
& A-GEM~\cite{ICLR2019_chaudhry} 	&	 22.67 $\pm$ 0.57 	&	 89.48 $\pm$ 1.45	&	8.06 $\pm$ 0.04 	&	 25.33 $\pm$ 0.49	& 80.31 $\pm$ 6.29	\\
& iCaRL~\cite{CVPR2017_Rebuffi} & 47.55 $\pm$ 3.95 & 88.22 $\pm$  2.62 & 9.38 $\pm$  1.53 & 31.55 $\pm$  3.27 & $*$ \\
& FDR~\cite{ICLR2019_Benjamin} 	&	 28.71 $\pm$ 3.23 	&	 93.29 $\pm$ 0.59 	&	 10.54 $\pm$ 0.21 	&	 49.88 $\pm$ 0.71 	& 89.67	 $\pm$ 1.63 \\
& DER++~\cite{NIPS2020_Buzzega} 	&	 72.70 $\pm$ 1.36 	&	 93.88 $\pm$ 0.50 	&	 19.38 $\pm$ 1.41 	&	 51.91 $\pm$ 0.68 	& 92.77 $\pm$ 1.05	\\
& Co2L~\cite{ICCV2021_Cha} 	&	 74.26 $\pm$ 0.77 	&	{\bf 95.90} $\pm$ 0.26 	&	 20.12 $\pm$ 0.42 	&	 53.04 $\pm$ 0.69 	& -	\\
& TARC~\cite{bhat2022task} 	&	 67.41 $\pm$ 0.41 	&	 - 	&	 13.77 $\pm$ 0.17 	&	 - 	& -	\\
& ER-ACE~\cite{ICLR2022_Caccia} 	&	 68.45 $\pm$ 1.78 	&	 93.47 $\pm$ 1.00 	&	 17.73 $\pm$ 0.56 	&	 49.99 $\pm$ 1.51 	& -	\\
& DRI~\cite{AAAI2022_Wang} 	&	 72.78 $\pm$ 1.44 	&	 93.85 $\pm$ 0.46 	&	 22.63 $\pm$ 0.81 	&	 52.89 $\pm$ 0.60 	& 93.02 $\pm$ 0.85	\\
& TAMiL~\cite{ICLR2023_Bhat} 	&	 74.45 $\pm$ 0.27 	&	 94.61 $\pm$ 0.19 	&	 28.48 $\pm$ 1.50 	&	 {\bf 65.19} $\pm$ 0.82 	&  $*$	\\
& SCoMMER~\cite{AAAI2023_Sarfraz} 	&	  74.97 $\pm$ 1.05 	&	 94.36  $\pm$ 0.06 	&	 - 	&	 - 	&	- \\ 
& CLS-ER~\cite{ICLR2022_AraniSZ} 	&	 {\bf 75.22} $\pm$ 0.71 	&	 94.35 $\pm$ 0.38 	& {\bf 29.61} $\pm$ 0.54 	&	 61.57 $\pm$ 0.63 	& {\bf 94.06} $\pm$ 0.07	\\
\midrule             
\multirow{12}{*}{200 samples}   
& ER~\cite{ICLR2019_Riemer} 	&	 44.79 $\pm$ 1.86 	&	 91.19 $\pm$ 0.94 	&	 8.57 $\pm$ 0.04 	&	 38.17 $\pm$ 2.00 	& 85.01 $\pm$ 1.90	\\
& A-GEM~\cite{ICLR2019_chaudhry} 	& 20.04 $\pm$ 0.34 	&	 83.88 $\pm$ 1.49	&	8.07 $\pm$ 0.08 	&	 22.77 $\pm$ 0.03	& 81.91 $\pm$ 0.76	\\
& iCaRL~\cite{CVPR2017_Rebuffi} &   49.02  $\pm$ 3.20 & 88.99 $\pm$ 2.13 & 7.53 $\pm$ 0.79 & 28.19 $\pm$ 1.47 & $*$ \\
& FDR~\cite{ICLR2019_Benjamin} 	&	 30.91 $\pm$ 2.74 	&	 91.01 $\pm$ 0.68 	&	 8.70 $\pm$ 0.19 	&	 40.36 $\pm$ 0.68 & 85.22 $\pm$ 3.55\\
& DER++~\cite{NIPS2020_Buzzega} 	&	 64.88 $\pm$ 1.17 	&	 91.92 $\pm$ 0.60 	&	 10.96 $\pm$ 1.17 	&	 40.87 $\pm$ 1.16 	& 90.43 $\pm$ 1.87	\\
& Co2L~\cite{ICCV2021_Cha} 	&	 65.57 $\pm$ 1.37 	&	 93.43 $\pm$ 0.78 	&	 13.88 $\pm$ 0.40 	&	 42.37 $\pm$ 0.74 	& -	\\
& TARC~\cite{bhat2022task} 	&	 53.23 $\pm$ 0.10 	&	 - 	&	 9.57 $\pm$ 0.12 	&	 - 	& -	\\
& ER-ACE~\cite{ICLR2022_Caccia} 	&	 62.08 $\pm$ 1.44 	&	 92.20 $\pm$ 0.57 	&	 11.25 $\pm$ 0.54 	&	 44.17 $\pm$ 1.02 	& -	\\
& DRI~\cite{AAAI2022_Wang} 	&	 65.16 $\pm$ 1.13 	&	 92.87 $\pm$ 0.71 	&	 17.58 $\pm$ 1.24 	&	 44.28 $\pm$ 1.37 	& 91.17 $\pm$ 1.53	\\
& TAMiL~\cite{ICLR2023_Bhat} 	&	 68.84 $\pm$ 1.18 	&	{\bf 94.28} $\pm$ 0.31 	&	 20.46 $\pm$ 0.40 	&	 55.44 $\pm$ 0.52 	& $*$	\\
& SCoMMER~\cite{AAAI2023_Sarfraz} 	&	 {\bf 69.19} $\pm$ 0.61 	&	 93.20  $\pm$ 0.10 	&	 - 	&	 - 	& -	 \\ 
& CLS-ER~\cite{ICLR2022_AraniSZ} 	&	 66.19 $\pm$ 0.75 	&	 93.59 $\pm$ 0.87 	&	{\bf 21.95} $\pm$ 0.26 	&	{\bf 58.41} $\pm$ 1.72 	& {\bf 92.26} $\pm$ 0.18	 \\
 \midrule
\multirow{2}{*}{Task ID}     
&	LwF~\cite{TPAMI2017_Li}	 &	19.61 $\pm$ 0.05	&	63.29 $\pm$ 2.35	&	8.46 $\pm$ 0.22	&	15.85 $\pm$ 0.58 &  $*$ \\
&	PNNs~\cite{Arxiv2016_Rusu}	&	-	&	95.13 $\pm$ 0.72	&	-	&	67.84 $\pm$ 0.29 	& $*$  \\  \midrule
\multirow{3}{*}{-}     
& oEWC~\cite{ICML2018_Schwarz}  & 19.49 $\pm$ 0.12 & 68.29 $\pm$ 3.92 &  7.58 $\pm$ 0.10 & 19.20 $\pm$ 0.31 & 77.35 $\pm$ 5.77 \\ 
& SI~\cite{ICML2017_Zenke}      & 19.48 $\pm$ 0.17 & 68.05 $\pm$ 5.91 &  6.58 $\pm$ 0.31 & 36.32 $\pm$ 0.13  & 71.91 $\pm$ 5.83 \\
% & BIR~\cite{NC2020_Ven}         & 46.14 $\pm$ 1.83 & 87.52 $\pm$ 0.91 &  - & -  & 73.26 $\pm$ 0.89 \\ 
& PFCL                          & {\bf 67.33} $\pm$ 0.54 & {\bf 96.13} $\pm$ 0.45 &  {\bf 18.75} $\pm$ 0.16 &  {\bf 69.70} $\pm$ 0.56 & {\bf 82.58} $\pm$ 0.73 \\ \bottomrule     
\end{tabular}}
\label{tbl_all}
\end{table*}    

\subsection{Comparison Results}
Following the experimental settings in~\cite{NIPS2020_Buzzega} and~\cite{ICLR2023_Bhat}, we split each dataset (CIFAR10, CIFAR100, and Tiny ImageNet) into sequence tasks of disjoint classes to evaluate the performance of Class-IL and Task-IL. Specifically, we evaluate our model on CIFAR10 with 5 tasks (CIFAR10-5)~\cite{NIPS2020_Buzzega}, CIFAR100 with multiple lengths of tasks (including 5, 10, and 20)~\cite{CVPR2022_Wang_vit}, and Tiny ImageNet with 10 tasks (TinyImg-10)~\cite{NIPS2020_Buzzega}. Additionally, we evaluate the performance of Domain-IL on the Rotated MNIST (RMNIST-20) dataset with 20 tasks. Similar to most CL methods, we mainly discuss the performance of the proposed method in Class-IL.

{\bf Baselines.} For fairness, we compare the proposed PFCL with several methods that use the same backbone. Because of different problem settings, we do not compare the methods discussed in Section~\ref{sec_unlabeled}. To show the effectiveness of these CL methods, we provide a lower bound method denoted as FT by simply fine-tuning and an upper bound method denoted as the JT by jointly training all tasks offline. Besides, we divide the baselines into three groups according to the usage of different task priors during model training.
\begin{enumerate}
    \item {\bf Previous samples.} We report the performance of several rehearsal-based methods for comparisons, including ER~\cite{ICLR2019_Riemer}, A-GEM~\cite{ICLR2019_chaudhry}, iCaRL~\cite{CVPR2017_Rebuffi}, FDR~\cite{ICLR2019_Benjamin}, DER++~\cite{NIPS2020_Buzzega}, Co2L~\cite{ICCV2021_Cha}, TARC~\cite{bhat2022task}, ER-ACE~\cite{ICLR2022_Caccia}, DRI~\cite{AAAI2022_Wang}, TAMiL~\cite{ICLR2023_Bhat}, SCoMMER~\cite{AAAI2023_Sarfraz}, CLS-ER~\cite{ICLR2022_AraniSZ}, HAL~\cite{AAAI2021_Chaudhry}, ERT~\cite{ICPR2021_Buzzega}, and RM~\cite{CVPR2021_Bang}. Among these approaches, iCaRL and TAMiL require task identity to learn task-specific knowledge, and thus they cannot be applied in Domain-IL scenarios. DRI is a generative method that produces synthetic data with GAN. TAMiL, SCoMMER, and CLS-ER require two complementary learning systems to balance stability and plasticity. Following the recent methods~\cite{ICLR2022_AraniSZ, AAAI2023_Sarfraz, ICLR2023_Bhat}, we report the performance of all rehearsal-based methods with popularly used memory buffer sizes of 200 and 500. 
    
    \item {\bf Task ID.} We compare two rehearsal-free CL methods that reduce forgetting with task identity only. PNN~\cite{Arxiv2016_Rusu} is a dynamic architecture-based method, it requires task identity during both training and inference times. Following DER++~\cite{TPAMI2017_Li}, LwF used in our experiments employs a single-head architecture and requires task identity during the training stage only. LwF~\cite{TPAMI2017_Li} uses knowledge distillation in prediction spaces and stores the old model's responses to the new task at the beginning of each task. Compared to LwF, our PFCL does not use task identity or store data during training.
    
    \item {\bf Prior-Free.} Reducing forgetting without any task prior is a very general and challenging task, but prior-free CL is seldom studied. We compare with two regularization-based methods oEWC~\cite{ICML2018_Schwarz} and SI~\cite{ICML2017_Zenke} that seek model consistency in parameter spaces. Compared to these two methods, the proposed PFCL method is more straightforward and we seek model consistency in prediction spaces. In addition, PFCL leverages auxiliary unlabeled data to assist model regularization additionally.
\end{enumerate}

% \begin{table*}[]
% \centering
% \caption{Classification results of PFCL with different losses. KL denotes the Kullback-Leibler divergence loss and MSE represents the Mean Squared Error loss.}
% \resizebox{\linewidth}{!}{
% \begin{tabular}{lccccccccccc}
% \toprule
% \multirow{2}{*}{Loss} & \multicolumn{2}{c}{CIFAR10-5} & \multicolumn{2}{c}{CIFAR100-5} & \multicolumn{2}{c}{CIFAR100-10} & \multicolumn{2}{c}{CIFAR100-20} & \multicolumn{2}{c}{TinyImg-10} & RMNIST-20 \\  
% & Class-IL    & Task-IL   & Class-IL    & Task-IL   & Class-IL    & Task-IL   & Class-IL   & Task-IL  & Class-IL  & Task-IL   & Domain-IL \\ \midrule
% KL & {\bf 67.33}  &  96.13    &   42.86  & 81.08   &  {\bf 29.83}   &  84.32  & {\bf 21.22}  &  {\bf 84.29}  & {\bf 18.64}  &  69.73  & {\bf 82.58}   \\
% MSE    \\ \bottomrule
% \end{tabular}}
% \label{tbl_loss}
% \end{table*}

\begin{table*}[t]
\centering
\caption{Classification results of Class-IL and Task-IL on CIFAR100 benchmark dataset with a different number of tasks, averaged across 3 runs.}
\resizebox{0.95\linewidth}{!}{
\begin{tabular}{clcccccc}
\toprule
\multirow{2}{*}{\begin{tabular}[c]{@{}l@{}}Prior  \end{tabular}} & \multirow{2}{*}{Method} & \multicolumn{2}{c}{CIFAR100-5}  & \multicolumn{2}{c}{CIFAR100-10} & \multicolumn{2}{c}{CIFAR100-20}    \\ 
&   & Class-IL    & Task-IL      & Class-IL     & Task-IL      & Class-IL & Task-IL \\ \midrule    
\multirow{2}{*}{-}  
& JT (upper bound) & 70.21 $\pm$ 0.15 & 85.25 $\pm$ 0.29 & 70.21 $\pm$ 0.15 & 91.24 $\pm$ 0.27 & 71.25 $\pm$ 0.22 & 94.02 $\pm$ 0.33 \\
& FT (lower bound)   & 17.27 $\pm$ 0.14 & 42.24 $\pm$ 0.33 & 8.62 $\pm$ 0.09 & 34.40 $\pm$ 0.53 & 4.73 $\pm$ 0.06 & 40.83 $\pm$ 0.46   \\  \midrule  	   
\multirow{12}{*}{500 samples} 
& ER~\cite{ICLR2019_Riemer}  & 27.97 $\pm$ 0.33 & 68.21 $\pm$ 0.29 & 21.54 $\pm$ 0.29 & 74.97 $\pm$ 0.41 & 15.36 $\pm$ 1.15 & 74.97 $\pm$ 1.44    \\ 
% & GEM~\cite{NIPS2017_Lopez} & 25.44 $\pm$ 0.72 & 67.49 $\pm$ 0.91 & 18.48 $\pm$ 1.34 & 72.68 $\pm$ 0.46 & 12.58 $\pm$ 2.15 & 78.24 $\pm$ 0.61   \\ 
& A-GEM~\cite{ICLR2019_chaudhry}  & 18.75 $\pm$ 0.51 & 58.70 $\pm$ 1.49 & 9.72 $\pm$ 0.22 & 58.23 $\pm$ 0.64 & 5.97 $\pm$ 1.13 & 59.12 $\pm$ 1.57         \\ 
& iCaRL~\cite{CVPR2017_Rebuffi} & 35.95 $\pm$ 2.16 & 64.40 $\pm$ 1.59 & 30.25 $\pm$ 1.86 & 71.02 $\pm$ 2.54 & 20.05 $\pm$ 1.33 & 72.26 $\pm$ 1.47 \\ 
% & GSS~\cite{NIPS2019_Aljundi}   & 22.08 $\pm$ 3.51 & 61.77 $\pm$ 1.52 & 13.72 $\pm$ 2.64 & 56.32 $\pm$ 1.84 & 7.49 $\pm$ 4.78 & 57.42 $\pm$ 1.61    \\
& FDR~\cite{ICLR2019_Benjamin}   & 29.99 $\pm$ 2.23 & 69.11 $\pm$ 0.59 & 22.81 $\pm$ 2.81 & 74.22 $\pm$ 0.72 & 13.10 $\pm$ 3.34 & 73.22 $\pm$ 0.83   \\ 
& HAL~\cite{AAAI2021_Chaudhry}   & 16.74 $\pm$ 3.51 & 39.70 $\pm$ 2.53 & 11.12 $\pm$ 3.80 & 41.75 $\pm$ 2.17 & 9.71 $\pm$ 2.91 & 55.60 $\pm$ 1.83    \\ 	      
& DER++~\cite{NIPS2020_Buzzega} & 38.39 $\pm$ 1.57 & 70.74 $\pm$ 0.56 & 36.15 $\pm$ 1.10 & 73.31 $\pm$ 0.78 & 21.65 $\pm$ 1.44 & 76.55 $\pm$ 0.87 \\ 
& ERT~\cite{ICPR2021_Buzzega}   & 28.82 $\pm$ 1.83 & 62.85 $\pm$ 0.28 & 23.00 $\pm$ 0.58 & 68.26 $\pm$ 0.83 & 18.42 $\pm$ 1.92 & 73.50 $\pm$ 0.82   \\ 
& RM~\cite{CVPR2021_Bang}  & 39.47 $\pm$ 1.26 & 69.27 $\pm$ 0.41 & 32.52 $\pm$ 1.53 & 73.51 $\pm$ 0.89 & 23.09 $\pm$ 1.72 & 75.06 $\pm$ 0.75    \\ 
& ER-ACE~\cite{ICLR2022_Caccia}	&	40.67 $\pm$ 0.06	&	66.45 $\pm$ 0.71	& 36.18 $\pm$ 1.44	&	74.70 $\pm$ 0.57	&	30.72 $\pm$ 1.17	&	79.59 $\pm$ 1.23	\\
& TAMiL~\cite{ICLR2023_Bhat} & {\bf 50.11} $\pm$ 0.34 & {\bf 76.38} $\pm$ 0.30 & {\bf 44.48} $\pm$ 1.18 & {\bf 80.43} $\pm$ 0.31 & 29.35 $\pm$ 0.75 & 79.70 $\pm$ 0.17	\\  
& SCoMMER~\cite{AAAI2023_Sarfraz} & 49.63  $\pm$ 1.43	& 73.49  $\pm$ 0.43 &  35.89 $\pm$ 0.61 & 78.70	 $\pm$ 0.52	&	29.75  $\pm$ 0.35	&	81.09  $\pm$ 0.43	\\  
& CLS-ER~\cite{ICLR2022_AraniSZ} & 47.63 $\pm$ 0.61 & 73.78 $\pm$ 0.86 &	43.12 $\pm$ 0.75 &	78.59 $\pm$ 0.87 &	{\bf 34.59} $\pm$ 0.86	&	{\bf 81.74} $\pm$ 0.79	\\  \midrule             
\multirow{12}{*}{200 samples}   
& ER~\cite{ICLR2019_Riemer}  & 21.94 $\pm$ 0.83 & 62.41 $\pm$ 0.93 & 14.23 $\pm$ 0.12 & 62.57 $\pm$ 0.68 & 9.90 $\pm$ 1.67 & 70.82 $\pm$ 0.74       \\
% & GEM~\cite{NIPS2017_Lopez} & 19.73 $\pm$ 0.34 & 57.13 $\pm$ 0.94 & 13.20 $\pm$ 0.21 & 62.96 $\pm$ 0.67 & 8.29 $\pm$ 0.18 & 66.28 $\pm$ 1.49      \\
& A-GEM~\cite{ICLR2019_chaudhry} & 17.97 $\pm$ 0.26 & 53.55 $\pm$ 1.13 & 9.44 $\pm$ 0.29 & 55.04 $\pm$ 0.87 & 4.88 $\pm$ 0.09 & 41.30 $\pm$ 0.56      \\
& iCaRL~\cite{CVPR2017_Rebuffi} & 30.12 $\pm$ 2.45 & 55.70 $\pm$ 1.87 & 22.38 $\pm$ 2.79 & 60.81 $\pm$ 2.48 & 12.62 $\pm$ 1.43 & 62.17 $\pm$ 1.93   \\
% & GSS~\cite{NIPS2019_Aljundi} & 19.44 $\pm$ 2.83 & 56.11 $\pm$ 1.50 & 11.84 $\pm$ 1.46 & 56.24 $\pm$ 0.98 & 6.42 $\pm$ 1.24 & 51.64 $\pm$ 2.89   \\
& FDR~\cite{ICLR2019_Benjamin} & 22.84 $\pm$ 1.49 & 63.75 $\pm$ 0.49 & 14.85 $\pm$ 2.76 & 65.88 $\pm$ 0.60 & 6.70 $\pm$ 0.79 & 59.13 $\pm$ 0.73      \\
& HAL~\cite{AAAI2021_Chaudhry} & 13.21 $\pm$ 1.24 & 35.61 $\pm$ 2.95 & 9.67 $\pm$ 1.67 & 37.49 $\pm$ 2.16 & 5.67 $\pm$ 0.91 & 53.06 $\pm$ 2.87     \\
& DER++~\cite{NIPS2020_Buzzega} & 27.46 $\pm$ 1.16 & 62.55 $\pm$ 2.31 & 21.76 $\pm$ 0.78 & 63.54 $\pm$ 0.77 & 15.16 $\pm$ 1.53 & 71.28 $\pm$ 0.91    \\
& ERT~\cite{ICPR2021_Buzzega} & 21.61 $\pm$ 0.87 & 54.75 $\pm$ 1.32 & 12.91 $\pm$ 1.46 & 58.49 $\pm$ 3.12 & 10.14 $\pm$ 1.96 & 62.90 $\pm$ 2.72 \\   
& RM~\cite{CVPR2021_Bang} & 32.23 $\pm$ 1.09 & 62.05 $\pm$ 0.62 & 22.71 $\pm$ 0.93 & 66.28 $\pm$ 0.60 & 15.15 $\pm$ 2.14 & 68.21 $\pm$ 0.43 \\ 
& ER-ACE~\cite{ICLR2022_Caccia}	& 35.17 $\pm$ 1.17	& 63.09 $\pm$ 1.23 & 27.68 $\pm$ 1.24 &	68.68 $\pm$ 0.52 & 21.17 $\pm$ 1.17 & 77.29 $\pm$ 1.43	\\
& TAMiL~\cite{ICLR2023_Bhat} & {\bf 41.43} $\pm$ 0.75 &	{\bf 71.39} $\pm$ 0.17&	32.23 $\pm$ 1.18 &	{\bf 74.62} $\pm$ 0.31	& 19.20 $\pm$ 0.75 & 74.42 $\pm$ 0.17	\\  
& SCoMMER~\cite{AAAI2023_Sarfraz} &	40.25  $\pm$ 0.05	&	69.39  $\pm$ 0.43	&  22.89 $\pm$ 0.61 & 70.53	 $\pm$ 0.10	&	19.25  $\pm$ 0.05	&	76.79  $\pm$ 0.43  \\ 
& CLS-ER~\cite{ICLR2022_AraniSZ} &	35.23 $\pm$ 0.86 &	67.34 $\pm$ 0.79 & {\bf 32.55} $\pm$ 0.75 & 71.42 $\pm$ 0.87 & {\bf 25.23} $\pm$ 0.86 & {\bf 77.34} $\pm$ 0.79	 \\ \midrule          
\multirow{1}{*}{Task ID}  
& LwF~\cite{TPAMI2017_Li}    & 18.16 $\pm$ 0.18 & 30.61 $\pm$ 1.49 & 9.41 $\pm$ 0.06 & 28.69 $\pm$ 0.34  & 4.82 $\pm$ 0.06 & 39.38 $\pm$ 1.10 \\   \midrule  
\multirow{3}{*}{-}   
& oEWC~\cite{ICML2018_Schwarz} & 16.92 $\pm$ 0.28 & 31.51 $\pm$ 1.02 & 8.11 $\pm$ 0.47 & 23.21 $\pm$ 0.49  & 4.44 $\pm$ 0.17 & 26.48 $\pm$ 2.07 \\ 
& SI~\cite{ICML2017_Zenke}     & 17.60 $\pm$ 0.09 & 43.64 $\pm$ 1.11 & 9.39 $\pm$ 0.61 & 29.32 $\pm$ 2.03  & 4.47 $\pm$ 0.07 & 32.53 $\pm$ 2.70 \\  
& PFCL    & {\bf 42.86} $\pm$ 0.39 & {\bf 81.08} $\pm$ 0.61 & {\bf 29.83} $\pm$ 0.42 & {\bf 84.32} $\pm$ 0.23  & {\bf 21.22} $\pm$ 0.70  & {\bf 84.29} $\pm$ 0.35 \\ \bottomrule    
\end{tabular}}
\label{tbl_cifar100}
\end{table*}           

\begin{figure*}[ht]
\centering
\begin{subfigure}{0.47\linewidth}
  \centering
  \includegraphics[width=\linewidth]{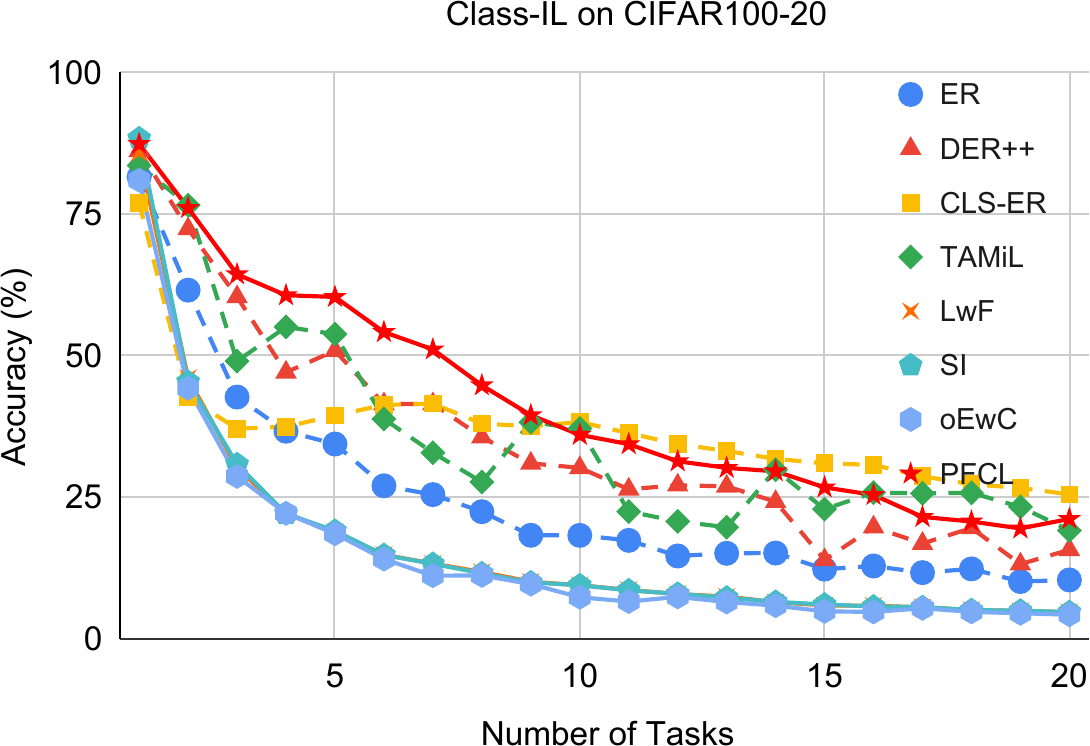}  
\end{subfigure}
\hspace{1em}
\begin{subfigure}{.48\linewidth}
  \centering
  \includegraphics[width=\linewidth]{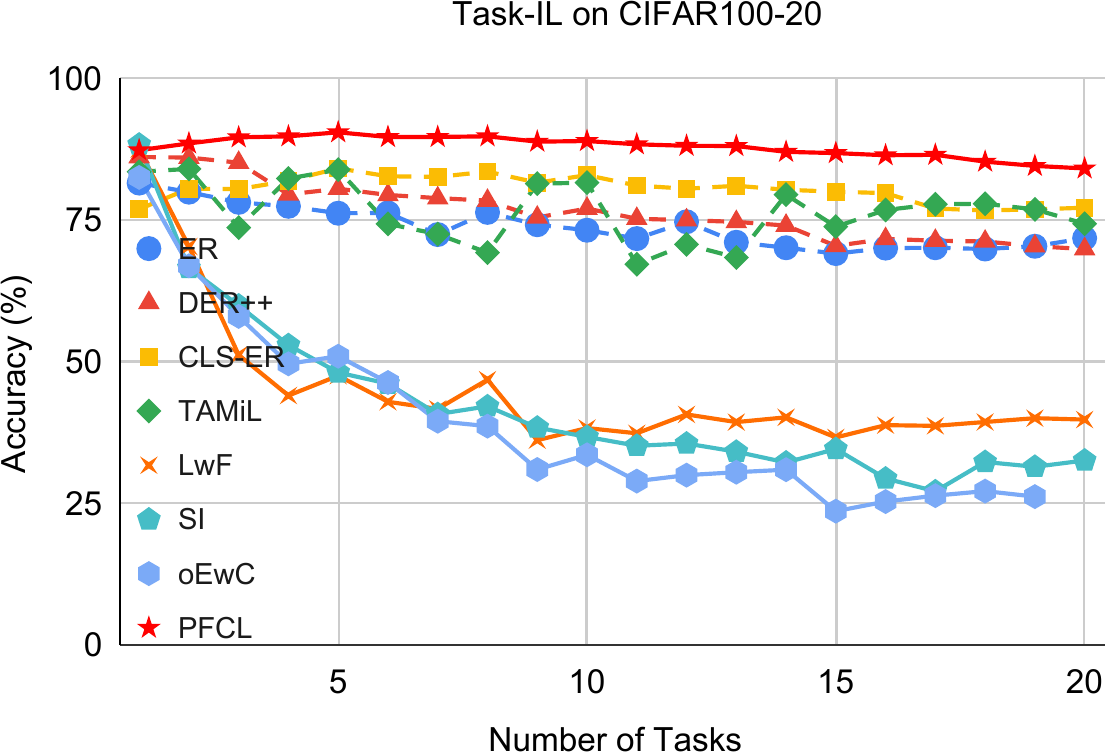}  
  % \label{fig:sub-first}
\end{subfigure}
\caption{Average accuracy of Class-IL and Task-IL when incrementally learning 20 tasks on CIFAR100 dataset. The memory size of rehearsal-based methods is 200.}
\label{fig_cifar100}
\end{figure*}

{\bf Overall Performance.}
\tab~\ref{tbl_all} presents the average accuracy of PFCL in all three CL scenarios. It can be seen that PFCL significantly alleviates the forgetting issue when compared to FT. Even compared to the most recent rehearsal-based methods (\eg TAMiL, SCoMMER, and CLS-ER) that replay 200 samples in Class-IL, the performance of PFCL is competitive. Moreover, PFCL outperforms all compared methods in Task-IL. Without revisiting any previous samples, PFCL surpasses all rehearsal-free techniques by a large margin in all experiments, verifying the effectiveness of our method. Detailed discussions are as follows.

{\bf Results of Class-IL.} 
Class-IL sequentially learns new classes without requiring task identity at the reference time. Existing regularization-based methods have a major drawback: inability in Class-IL, especially for a long stream of tasks. The proposed PFCL method overcomes this problem well. \tab~\ref{tbl_all} shows the average accuracy of CL methods after learning all tasks on CIFAR10 and TinyImageNet. We can observe that the accuracy of previous prior-free methods is poor. For example, the performance of LwF, oEWC, and SI is close to the lower bound FT on all experiments, which indicates that these methods fail to retain knowledge in Class-IL. Such results have been also observed in previous methods~\cite{NIPS2020_Buzzega, NeuComp2021}. By seeking consistent predictions with an auxiliary dataset additionally, PFCL outperforms the rehearsal-free methods by a large margin, \eg 47.7\% on CIFAR10-5 and 11.2\% on TinyImg-10. \tab~\ref{tbl_cifar100} presents the results on the CIFAR100 dataset with multiple lengths of tasks. It can be seen that PFCL surpasses the rehearsal-free based methods by a large margin, such as 20.4\% on CIFAR100-10 and 16.4\% on CIFAR100-20. Such results show that the proposed PFCL method effectively reduces forgetting for a large number of tasks. 

By replaying a subset of previous samples, rehearsal-based approaches usually obtain better results than rehearsal-free methods. However, the performance of rehearsal-based methods heavily depends on the number of available samples, both \tab~\ref{tbl_all} and \tab~\ref{tbl_cifar100} also confirm this weakness. Without using any task priors, PFCL outperforms many rehearsal-based methods in all experiments when 500 previous samples are provided for them, \eg ER, A-GEM, and FDR. In particular, PFCL achieves comparable accuracy to several most recent methods when they replay 200 samples, such as TAMiL, SCoMMER, and CLS-ER. Although PFCL leverages more data than the rehearsal-based methods in our experiments, notice that the auxiliary data does not contain any task prior and it can be freely collected in the wild. Therefore, the problem setting of our method is more challenging than rehearsal-based methods.

\fig~\ref{fig_cifar100} (left image) shows the average accuracy of different methods when incrementally learning 20 tasks on CIFAR100. PFCL outperforms the previous rehearsal-free methods by a large margin after each learning step, verifying the effectiveness of our method. Moreover, PFCL surpasses the rehearsal-based methods before 10 tasks and achieves comparable accuracy after learning all 20 tasks. Based on the above observations, it can be concluded that the proposed PFCL method effectively overcomes the major drawback of existing regularization-based methods in Class-IL.

\begin{table*}[!t]
\centering
\caption{Forgetting results of rehearsal-free CL methods in Class-IL and Task-IL (lower is better).}
\resizebox{\linewidth}{!}{
\begin{tabular}{clcccccccccc}
\toprule
\multirow{2}{*}{Prior} & \multirow{2}{*}{Method} & \multicolumn{2}{c}{CIFAR10-5} & \multicolumn{2}{c}{CIFAR100-5} & \multicolumn{2}{c}{CIFAR100-10} & \multicolumn{2}{c}{CIFAR100-20} & \multicolumn{2}{c}{TinyImg-10} \\  
& & Class-IL    & Task-IL    & Class-IL    & Task-IL    & Class-IL   & Task-IL   & Class-IL    & Task-IL    & Class-IL   & Task-IL   \\ \midrule 
\multirow{1}{*}{Task ID}  
& LwF~\cite{TPAMI2017_Li}   & 96.69 & 32.56  & 83.41 & 68.30 & 89.94  & 67.30 & 92.26 & 57.73 & 76.35 & 67.23 \\   \midrule  
\multirow{3}{*}{-}   
& oEWC~\cite{ICML2018_Schwarz} & 91.64 & 29.33 & 79.96 & 63.35 & 83.18 & 71.15 & 79.26 & 61.43 & 73.66 & 63.02  \\ 
& SI~\cite{ICML2017_Zenke}   & 95.78 & 38.76 & 83.54 & 50.28 & 88.48 & 63.43 & 92.07 & 66.68 & 67.61 & 33.60 \\  
& PFCL     & {\bf 23.86} & {\bf 1.22} & {\bf 29.85} & {\bf 5.34} & {\bf 51.32} & {\bf 5.96} & {\bf 65.35} & {\bf 9.62} & {\bf 47.16} & {\bf 9.63} \\ \bottomrule 
\end{tabular}}
\label{tbl_forget}
\end{table*}

\begin{table*}[t]
\centering
\caption{Classification results of PFCL with different modules. RSS denotes the reliable sample selection module.}
\resizebox{\linewidth}{!}{
\begin{tabular}{lccccccccccc}
\toprule
\multirow{2}{*}{Module} & \multicolumn{2}{c}{CIFAR10-5} & \multicolumn{2}{c}{CIFAR100-5} & \multicolumn{2}{c}{CIFAR100-10} & \multicolumn{2}{c}{CIFAR100-20} & \multicolumn{2}{c}{TinyImg-10} & RMNIST-20 \\  
                  & Class-IL       & Task-IL      & Class-IL       & Task-IL       & Class-IL        & Task-IL       & Class-IL        & Task-IL       & Class-IL       & Task-IL       & Domain-IL \\ \midrule
$\mathcal{D}_t$    & 20.71  & 94.68   &  30.39  &  78.54  & 12.92  &  75.64  & 6.54  & 66.26 & {\bf 19.16}  & 69.70  &  80.49        \\
$\mathcal{D}_u$    & 25.93  & 74.27   &  20.72  &  47.16  & 11.52  &  33.90  & 5.94  & 35.95 & 7.91   & 23.46  &  80.74         \\
$\mathcal{D}_t \cup \mathcal{D}_u$   & 60.88  & {\bf 96.39}  &  42.71   &  {\bf 81.49}   & 28.88  &  83.24  &  19.32   &  83.35   &  14.84  &  67.26   &   80.50        \\
$\mathcal{D}_t \cup \mathcal{D}_u$+RSS    &  {\bf 67.33}  &  96.13    &   {\bf 42.86}  & 81.28   &  {\bf 29.83}   &  {\bf 84.32}   & {\bf 21.22}  &  {\bf 84.29}  & 18.75    &   {\bf 69.73}   &  {\bf 82.58}    \\  \bottomrule
\end{tabular}}
\label{tbl_moduels}
\end{table*}

{\bf Results of Task-IL.} 
Following previous methods~\cite{NIPS2020_Buzzega}, we do not use task identity for model training in all experiments, including the Task-IL scenarios. Based on the same predictions of Class-IL, we evaluate the performance of Task-IL by using a given task identity to select task-specific output units. \tab~\ref{tbl_all} and \tab~\ref{tbl_cifar100} show that PFCL outperforms all compared methods in all experiments, even including the Task-IL method PNN. Especially, PFCL surpasses the rehearsal-free methods by a large margin of 55\% on CIFAR100-10 and 45\% on CIFAR100-20. In addition, compared to the rehearsal-based methods with 500 samples, PFCL also obtains better performance. Different from LwF which employs task identities to select task-specific outputs during training, we directly use the full output spaces for regularization. Therefore, PFCL can effectively preserve the learned knowledge in a larger space and well adapt the model to multiple tasks when a task identity is provided. 

{\bf Results of Domain-IL.} 
Domain-IL aims to learn streaming data with different domain shifts, where task identity is unknown at all times. Therefore, many CL methods cannot be applied in this scenario, such as the typical regularization-based method LwF and some rehearsal-based methods (iCaRL and TAMiL). \tab~\ref{tbl_all} shows the comparison results of Domain-IL on Rotated MNIST (20 tasks). It can be seen that PFCL achieves the best performance among the rehearsal-free methods and it obtains noticeable performance improvement. Since the class labels between tasks are the same in Domain-IL, using a regularization-based alone may overwrite the previous knowledge and lead to forgetting. By contrast, rehearsal-based methods obtain better performance by replaying a few samples.

{\bf Forgetting.} \tab~\ref{tbl_forget} reports the forgetting results of rehearsal-free methods in Class-IL and Task-IL. By incorporating an auxiliary dataset to assist model regularization and a reliable sample selection to enhance consistency, PFCL outperforms the compared approaches by a large margin in both Class-IL and Task-IL. These results indicate that PFCL effectively mitigates forgetting.

% {\bf Loss functions.} 
% In continual learning, the training loss for knowledge distillation is computed by KL divergence~\cite{TPAMI2017_Li} or MSE loss~\cite{NIPS2020_Buzzega, ICLR2022_AraniSZ}. Unlike KL divergence loss, MSE loss directly matches the logits of the teacher model without softening, which is denoted as:
% \begin{equation}
%      \ell_{mse} = \|f(x; \theta_n) - f(x; \theta_o)\|^2.
%     \label{eq_mse}
% \end{equation}
% \tab~\ref{tbl_loss} shows the classification results of the proposed method on all tasks with different losses. It can be seen that KL divergence loss outperforms MSE loss by a large margin. This is because auxiliary data does not provide accurate information on old tasks. 
%MSE may transfer the negative information to the new model. In our experiments, we mainly report the performance with KL divergence loss.

\subsection{Ablation Study and Analysis}
Using a conventional regularization-based method alone usually fails to reduce forgetting in Class-IL. To address this issue, PFCL leverages auxiliary unlabeled data to assist model regularization and designs a reliable sample selection to seek consistent performance improvement. To demonstrate the effectiveness of each module, we report the results of different modules on multiple datasets.

{\bf Auxiliary unlabeled data.} \tab~\ref{tbl_moduels} shows that using the current training data alone significantly reduces forgetting in Task-IL, but it fails in Class-IL. On the other hand, using the auxiliary dataset alone fails to maintain acquired knowledge in both Class-IL and Task-IL. The reason is that the auxiliary dataset does not provide past information during training, and seeking consistent predictions on it cannot retain task-specific knowledge effectively. By seeking model consistency with two datasets together, the average accuracy is greatly improved in Class-IL and slightly boosted in Domain-IL. 

{\bf Reliable sample selection.} Directly using all data for model regularization may degrade the performance, such as the accuracy of TinyImg-10, we design a reliable sample selection method to solve this issue. Although the accuracy of Task-IL slightly drops on CIFAR10-5 and CIFAR100-5, it can be seen that the overall performance is further improved in both Class-IL and Task-IL. In Domain-IL, we can also observe consistent performance improvement by using auxiliary unlabeled data and the reliable sample selection module. 

For TinyImg-10 with 200 classes, using auxiliary data and reliable sample selection does not obtain performance improvement, this is because the performance of the proposed method relies on the auxiliary datasets. Next, we discuss the effects of the auxiliary dataset as follows. 

\begin{table*}[t]
\centering
\caption{Classification results of PFCL with different auxiliary datasets. }
\resizebox{\linewidth}{!}{
\begin{tabular}{lcccccccccccc}
\toprule
\multirow{2}{*}{Dataset} & \multirow{2}{*}{Size} & \multicolumn{2}{c}{CIFAR10-5} & \multicolumn{2}{c}{CIFAR100-5} & \multicolumn{2}{c}{CIFAR100-10} & \multicolumn{2}{c}{CIFAR100-20} & \multicolumn{2}{c}{TinyImg-10} & RMNIST-20 \\  
                  & & Class-IL       & Task-IL      & Class-IL       & Task-IL       & Class-IL        & Task-IL       & Class-IL        & Task-IL       & Class-IL       & Task-IL       & Domain-IL \\ \midrule
Flowers102  & 8,000 & 31.35  & 96.13   &  37.00  &  80.34  & 16.85  &  78.60  & 10.19  & 72.52 & 18.10  & 69.12  &  82.98        \\
XPIE\_N     & 8,000 & 49.07  & 96.22   &  38.19  &  80.65  & 21.64  &  79.12  & 14.62  & 78.81 & 18.62   & 69.81  &  84.77         \\
XPIE\_S     & 8,000 & {\bf 67.54}  & {\bf 96.57}  & 40.99   &  81.16   & 26.40  &  84.07  &  17.15   &  85.14   &  18.14  &  69.26   &  83.03   \\ \midrule
Caltech256  & 500 &  38.06  &  96.17 & 36.69  & 80.06 & 17.54   &  81.55   & 11.76  &  82.57  & 18.02    &   69.47   &  81.64  \\
Caltech256  & 5,000 & 65.28  &  96.53 & 41.32  & {\bf 81.46}   & 26.72    & 84.29    & 20.19  & 84.20  &  18.54   &  69.82    &  {\bf 85.76}   \\
Caltech256  & 8,000 & 66.35  &  96.16 & 41.79  & 81.31   &  27.72 &  84.05  & 19.67  & {\bf 85.20}  & 18.67  & {\bf 70.02}   &  85.41    \\  
Caltech256  & 30,607 & 67.33  &  96.13 & {\bf 42.86}  & 81.08 &  {\bf 29.83}   &  {\bf 84.32}   & {\bf 21.22}  &  84.29  & {\bf 18.75}  & 69.73 & 82.58   \\ \bottomrule
\end{tabular}}
\label{tbl_aux}
\end{table*}

\begin{figure*}[t]
\centering
\includegraphics[width=1.0\linewidth]{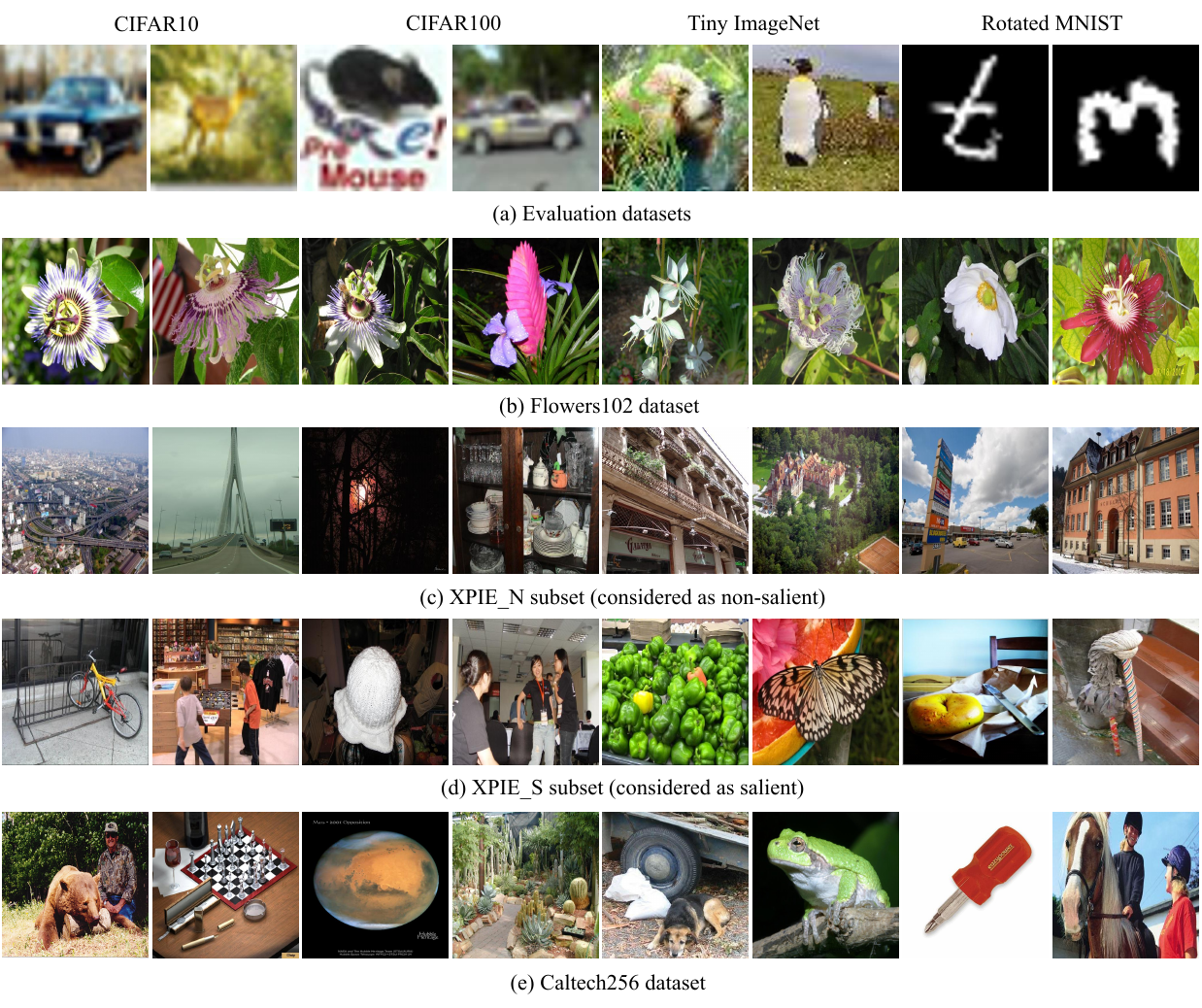}
\caption{Example images of the evaluation datasets and auxiliary datasets. The visual diversity of the Flowers102 dataset is lower than other auxiliary datasets because it consists of flower categories only.}
\label{fig_imgs}
\end{figure*}

\subsection{Effects of Different Auxiliary Datasets}
\label{sec_auxdata}

{\bf Auxiliary datasets.}
Auxiliary data plays an important role in our PFCL method. To evaluate the effects of different auxiliary datasets, we chose three datasets in our experiments. Detailed descriptions of these datasets are as follows.

{\bf Caltech256} dataset~\cite{Caltech256} consists of 30,607 real-world images and it spans 256 extremely diverse object categories and additional cluttered backgrounds, each class is represented by at least 80 images. Therefore, Caltech256 provides various data distributions for model regularization. 

{\bf Flowers102} dataset~\cite{Flowers102} has 8,189 images in 102 flower categories, each class consists of between 40 and 258 images. Because the flower images are similar to each other, the visual diversity of Flowers102 is lower than Caltech256.

{\bf XPIE} dataset~\cite{CVPR2017_Xia} was originally built for salient object detection. It contains 10,000 images containing salient objects (denoted as XPIE\_S) and 8,598 images without significant salient objects (denoted as XPIE\_N). Hence, XPIE\_N does not provide any object information of evaluation datasets. 

\fig~\ref{fig_imgs} presents some examples of the evaluation datasets and auxiliary datasets. It can be seen that their data distributions are very different. In this work, we mainly analyze the effect of auxiliary datasets from two aspects: visual diversity and dataset size. To analyze the visual diversity, we randomly select 8,000 images from each dataset for fair comparisons. Besides, to analyze the effect of data sizes, we randomly select 500, 5,000, and 8,000 samples from the Caltech256 dataset, respectively. 

{\bf Visual diversity.} \tab~\ref{tbl_aux} shows that the performance of PFCL relies on different auxiliary datasets. Given 8,000 auxiliary images for regularization, the overall performance of using Flower102 is worse than that of using other datasets in Class-IL and Task-IL. By contrast, using XPIE\_S and Caltech256 obtain better accuracy. In addition, although XPIE\_N does not contain any salient objects, it demonstrates better performance than Flower102 in Class-IL, especially on the CIFAR10-5 and CIFAR100-20. Compared to the results of Class-IL, the performance of Domain-IL does not heavily depend on different auxiliary datasets. These observations indicate that the auxiliary dataset has a significant impact on Class-IL. As we can see, Flower102 consists of flower images only, its visual diversity is lower than other datasets, and thus it does not greatly enhance model consistency. On the other hand, XPIE\_S and Caltech256 span diverse scenarios. Therefore, they can improve the consistency with rich data distributions and further retain acquired knowledge. 

{\bf Dataset size.} \tab~\ref{tbl_aux} reports the results of using different numbers of images. We can observe that selecting 5,000 images from Caltech256 greatly improves the classification results in Class-IL when compared to that of using 500 images, \eg from 38.06\% to 65.28\% on CIFAR10-5 and from 11.76\% to 20.19\% on CIFAR100-20. However, by increasing the number of images from 5,000 to 8,000, even to 30,607, the performance improvement is marginal. These findings imply that while auxiliary datasets can effectively reduce forgetting, they cannot completely retain learned knowledge through the use of additional images. As a result, prior-free continual learning is still a difficult issue to address.

\subsection{Model Discussion}
Based on extensive experiments, we discuss the advantages and limitations of the proposed PFCL method as follows.

{\bf Advantages.} (1) Unlike traditional CL methods, PFCL doesn’t require task identity or previous samples during training. This allows it to be applied in all three CL scenarios without knowing task priors. While an auxiliary unlabeled dataset is required, it can be freely collected in the wild and discarded after training, saving memory. (2) The performance of PFCL is competitive with recent rehearsal-based approaches that replay a limited number of samples. (3) PFCL primarily uses a knowledge distillation strategy, making it compatible with other CL techniques.

{\bf Limitations.} The performance of rehearsal-based methods could be consistently improved by storing more previous samples. By contrast, using a large number of auxiliary images in our method obtains marginal improvement. Despite using a reliable sample selection strategy, auxiliary data does not further boost the accuracy sometimes, such as on TinyImg-10. Because of distribution differences between auxiliary images and past samples, seeking prediction consistency on auxiliary data is insufficient for fully recovering previous knowledge. An effective sample generation approach without task priors may be a potential solution.

\section{Conclusion}
This paper introduces a simple and effective PFCL method that doesn’t require task identity or previous samples during training.  We first study the effectiveness and limitations of the conventional regularization-based method through extensive experiments. Then, we incorporate an auxiliary unlabeled dataset to enhance model consistency in prediction spaces and develop a reliable sample selection strategy to obtain consistent performance improvement. Extensive experiments on multiple image classification datasets show that the proposed PFCL method effectively reduces the forgetting issue in all three learning scenarios. Moreover, when a few past samples are available for rehearsal-based approaches, PFCL achieves comparable accuracy. We hope our study will inspire further research into the challenging but seldom-studied field of prior-free continual learning.

\bibliographystyle{IEEEtran}
\bibliography{egbib}

% Can use something like this to put references on a page
% by themselves when using endfloat and the captionsoff option.
% \ifCLASSOPTIONcaptionsoff
%   \newpage
% \fi

\end{document}